\documentclass[letterpaper, 10 pt, conference]{ieeeconf}  

\IEEEoverridecommandlockouts                              
\usepackage{amsmath}    
\usepackage{graphicx}   
\usepackage[font=footnotesize,labelfont=footnotesize]{caption}
\usepackage{verbatim}   
\usepackage{color}      
\usepackage{subfigure}  
\usepackage{hyperref} 
\usepackage{multirow}
\usepackage{rotating}
\usepackage{array}
\usepackage{xspace}
\usepackage{indentfirst}
\usepackage{amssymb}
\usepackage{float}
\usepackage{algorithm}
\usepackage{algorithmic}
\usepackage{comment}
\usepackage{sidecap}
\usepackage{booktabs}

\title{Robot Design: Formalisms, Representations, and the Role of the Designer}

\author{Alexandra Q. Nilles, Dylan A. Shell, and Jason M. O'Kane
\thanks{
{Alexandra Q. Nilles (\hbox{\texttt{nilles2@illinois.edu}}) is with the
Department of Computer Science at the University of Illinois, Urbana-Champaign,
Illinois, USA.}
{Dylan A. Shell (\hbox{\texttt{dshell@cse.tamu.edu}}) is
with the Department of Computer Science and Engineering, Texas A\&M University,
College Station, Texas, USA.}
{Jason M. O'Kane (\hbox{\texttt{jokane@cse.sc.edu}}) is with the Department
of Computer Science and Engineering, University of South Carolina, Columbia,
South Carolina, USA.} 
}}

\begin{document}

\maketitle

%

{\small
\begin{quotation}
``Civilization advances by extending the number of operations we can perform without thinking about them.''\newline \null\hfill --- Alfred North Whitehead 
\end{quotation}
}

\section{Introduction} 


Robotics is a-changin'.  In the very recent past, if a robot worked at all, one had cause to be happy.  But, moving beyond thinking merely about robots, some
researchers have begun to examine the robot design process itself.
We are beginning to see a broadening of scope from
the products to the process, the former stemming from expertise,
the latter being how that expertise is exercised.

The authors have participated in this discussion by helping to organize two
related workshops---the RSS 2016 Workshop on Minimality and Design Automation,
and the RSS 2017 Workshop on Minimality and Trade-offs in Automated Robot
Design. These workshops brought together researchers with a broad range of
specializations within robotics, including manipulation, locomotion, multi-robot
systems, bio-inspired robotics, and soft robotics; and who are developing lines
of research relevant to automated design, including formal methods, rapid
prototyping, discrete and continuous optimization, and development of new
software interfaces for robot design. Insights from these workshops heavily
inform the discussion we present here.\footnote{A full discussion of all the participants and research directions raised in the workshops lies beyond the scope of this abstract. See
\url{http://minimality.mit.edu/} for full speaker lists and more information.} This objective of this paper is to distill the following essential idea from those experiences:
\begin{quote}
	\sl The information abstractions popular within robotics, designed as they were to address insulated sub-problems, are currently inadequate for design automation. 
\end{quote}
To that end, this paper's first aim is to draw together multiple threads---specifically those
of formalization, minimality, automation, and integration---and to argue that
robot design questions involve some of the most interesting and fundamental
challenges for the discipline. While most efforts in
automating robot design have focused on optimization of hardware, robot design is
also inextricably linked to the design of the internal state of the robot, how that
internal state interacts with sensors and actuators, and how task specifications
are designed within this context. Focusing attention on those considerations is
worthwhile for the study of robot design because they are currently in a
critical intellectual sweet spot, being out of reach technically, but only just.

The second ingredient of this paper forms a roadmap. It emphasizes two aspects:
(1) the role of models in robot design, a reprise of the old chestnut about representation in robotics
(namely, that ``the world is its own best model''~\cite{brooks91iwr}); (2) a consideration of the human-element within the envisioned
scheme.


\section{Four Challenges for Design Automation}



From our experiences with robot design and the work toward automation of such,
four themes emerge which recur in many different problem spaces.

\subsection{Formalization: Toward Executable Robotics Theories}

Useful formalism builds symbolic models that enable chains of deduction in order
to make predictions and guarantees about robot performance. Robot
design problems exhibit a great deal of structure: we typically use a narrow set
of available hardware components, there are units for quantifying functionality
provided by components, and there are increasingly expressive languages for providing functional
specifications in classes of tasks~\cite{shoukry2016scalable,plaku2016motion}.

Once knowledge is abstracted in this way, it becomes
re-usable through the formation of libraries and tools. Models provide a way to
give expression to assumptions and guarantees; they are also the starting point
of a language of operations (such as composition, refinement, and compression)
to achieve higher levels of competency while managing complexity.
State-of-the-art techniques remain quite piecemeal, limited in the aspects of
the problems they encompass. Further, a great
deal of current knowledge is tied up in mathematical form, without being made
machine usable or readable---consider the criteria used to choose a particle
filter rather than an (extended) Kalman filter. If this expertise were encoded
formally, in
terms of model assumptions and resource trade-offs, software could provide a pose
estimate on the basis of domain properties without the roboticist being
concerned about the details.

Efforts to formalize robotics are moving toward more than ``on
paper" formalisms which capture the structure inherent in robotic systems. We
cannot ignore the work being done in formal synthesis techniques
\cite{kress2018synthesis} and related efforts to encode our knowledge about
robots and physical systems in an executable form. In turn, robotics provides a
plethora of benchmarks and motivating examples for researchers
in formal methods for embedded/hybrid/open systems,
beyond the classic applications such as thermostats and airplanes. In turn,
these researchers can benefit by engaging with prior work on formalizing robotics. The
results of these collaborations have major implications for the power and
correctness of automated design systems.

\subsection{Minimality: Toward an Understanding of Robot Power}



One decades-long line of research poses the question of what tasks a given robot can complete, or the
inverse question, what kinds of robots are capable of completing a given
task. If we imagine ranking robots by some
measure of complexity (their ability to sense, actuate, and compute), at the
bottom of this ranking is a robot with no sensors, no actuators, and no ability
to keep track of state. This robot is quite useless, except perhaps as a
paperweight. Then, as we augment the robot with sensors, actuators, and
computational power, at some point it becomes capable of accomplishing tasks.
The theoretical boundaries of this ``design space'' are not well understood,
despite considerable work in this direction \cite{blum1978power,donald1995information,OkaLav07b}.
As the robot design process becomes more automated, this line of
work becomes more relevant --- human designers may want to provide functional or
informational specifications (``I want a robot that can pick up this type of box''
or ``I want a robot that can find my keys in my living room'').

The theme of minimality is more than design-automation-through-optimization,
where we may ask how to design the smallest robot meeting some design
constraints, or one with the fewest number of linkages. This line of work is
useful, and advancing at an exciting pace (see, for example, the work of Spielberg \emph{et al.}~\cite{spielberg2017functional}
for an interesting example on simultaneous
optimization of robot design and motion strategy). However, it relies on
highly-trained humans to design the underlying models and design constraints. It
also does not help us explore the space of robots which may have quite different
body geometries and hardware designs, but which are all equivalent in their
power to complete a certain task. Thus, the theme of minimality intersects with
formalization, since better formalisms for describing and comparing the
functionality of robots are needed to reason about the theoretical limits of such systems. Progress in
this area would directly impact the power and correctness of automated design
tools.

Many of the problems in this space are computationally hard, in the sense of
NP-hardness \cite{o2017concise,ziglar2017context}. However, this is not a
reason to give up! Often, constraints on the problem space can bring design
problems back into a tractable realm, and advances in generic solvers for
problems like integer programming and satisfiability-modulo-theories (SMT) mean
that solving these problems is becoming more feasible in practice.

\subsection{Automation: Toward Tractable, Realizable Designs}


The prior two questions --- \emph{how} robotic systems can be computably represented,
and \emph{what} are the information requirements of robotic tasks --- are tied to
the question of automating the design and fabrication of robotic hardware.

One way to tackle the problem of representation is through modularity and
standardization. In the robot design space, standardization is largely driven by hardware
manufacturers. Robot designers generally, and academics or hobbyists especially, are
design robots by choosing from a range of off-the-shelf components. This constraint, along with work toward modular and composable robot
designs, helps tame the computational complexities involved.
When the information requirements of a given task have been identified, it
is easier to design a satisfactory robot from a finite collection of
components than to choose a design from the infinite space of all 
sensors, actuators, and body geometries.

Of course, the proliferation of new fabrication technologies is challenging this
view of the problem. Task satisfaction and hardware design algorithms are
becoming increasingly integrated, such as a system which automatically places winch-tendon networks in a soft robot based on a user-specified movement
profile~\cite{bern2017interactive}; or a system which evolves shapes of automatically fabricated wire
robots in order to achieve different specified locomotion tasks~\cite{cellucci20171d}; or a system which compiles high-level specifications
into laser-cut schematics and mechanical and wiring diagrams~\cite{mehta2015integrated}. The flexibility inherent in these approaches will require
new approaches to parameterizing and formalizing the very large design space
available to us.

\subsection{Integration: Toward an End-to-End System}

One of the largest challenges in robotics is the integration of different
components and control structures into one robotic system which functions
correctly, and ideally, has some guarantees on its performance. We must integrate 
mechanical, electrical, computational, and material systems, and also 
must reason across multiple levels of abstraction.

Similar to how computer systems have an abstraction ``stack'' (transistors to
byte code to programming languages to abstract reasoning over models of
programs), robotics has its own similar abstraction stack, which must give more
attention to the physical reality of the robotic system (from material
properties, to component implementations, to dataflow protocols, to abstract
sensor, actuator, and state representations, to task-level reasoning).\footnote{Note
that this ``design stack" has complicated dependencies between layers --- advances in low-level hardware enable
new choices at all levels of abstraction. This is true of traditional computer
architecture as well, though perhaps more true in robotics. The recent impact of
IMU availability on how robots are designed and programmed is a choice example.}

Identifying distinct layers of this abstraction stack, and the assumptions
therein, is a crucial challenge facing roboticists and is especially crucial for
those working on the frontiers of design automation. If this challenge is not
met, then high-level automation tools (such as formal logic specifications) which
aim to provide safety and security guarantees will be useless due to mismatches
with the physical implementations.

The difficulty of this integration task leads us to believe that robot design
will continue to be an iterative, experiment-driven process for the foreseeable
future, and tools which automate parts of the design process should enable this
workflow pattern. This is especially true wherever fabrication is a
time-consuming or otherwise expensive process. Even 3D printing, the consummate
rapid prototyping tool, can involve spending hours waiting for a print job to
finish - only to realize a flaw in the design when dynamics are taken into
account! The need for rapid feedback and prototyping is also great for 
robotic tasks which rely heavily on environmental interaction. Our current
simulation technology is not up to the task of determining if a given robot
design can navigate a sandy, rocky desert, for example.

\section{A tentative roadmap} 

{\small
\begin{quotation}
``Design activity... is a processes of `satisficing' rather than optimising;
producing any one of what might well be a large range of satisfactory solutions
rather than attempting to generate the one hypothetically-optimum solution."

--- Nigel Cross, \emph{Designerly Ways of Knowing} \cite{cross2001designerly}
\end{quotation}
}

\subsection{The Role of Models}

One frequent point of the discussion in the aforementioned RSS workshops on automated robot
design was the role of models in the robot design process. Some roboticists
argued that a "build first, model later (if at all)" approach is the most
effective method for robot design, and that automation efforts should focus on
making the prototyping process as fast as possible. Our lack of understanding of
the physics and ``unknown unknowns" in hardware implementations make most
models nearly useless in the design process. Even very high-fidelity simulations
often act completely differently than physical robots, and designer time is often
better spent making actual prototypes and observing their behavior.

The issue of what role models play in design also extends to optimization-based
approaches. This approach generally uses continuous and discrete optimization
algorithms to adjust robot morphology, sensor configurations, and motion
strategies. However, an algorithm which is optimizing the number and placements
of legs on a mobile robot will never spontaneously decide to try using wheels
instead. Often, the process of prototyping robots reveals constraints of the
task and available hardware that are not apparent at the beginning of the design
process. It is very unlikely that robot designers will create a perfect
optimization problem or other formal specification in a first attempt.  As a result, the design
process is inherently iterative, regardless of its degree of automation.

The following are design decisions (``forks in the road") that creators of robot
design tools can ask themselves to ensure thoughtful consideration of the models
used and their integration with the rest of the robot design ``stack":

\begin{itemize}
\item What assumptions does the tool make about what types of robots are being
designed?
\item Are modeling assumptions communicated clearly to users and (if
applicable) at the API level?
\begin{itemize}
\item For example, the Unified Robot Description Format (URDF), often used for
ROS robot models, only allows kinematic tree body types, and thus is unable to
specify robots with closed kinematic chains, but this assumption is clearly communicated in the documentation \cite{urdf}.
\end{itemize}
\item Are modeling assumptions enforced by the software (perhaps through type
systems, model checkers, a test set, etc), or does that responsibility fall to
the user? What kind of feedback does the tool give when these assumptions are
broken?
\item Does the tool attempt to give meaning to designs (such as visualization or
dynamical simulations) before they are fabricated?
\item How does the tool interact with the rest of the robot design ecosystem?
Can the tool leverage or bolster existing free and open-source technologies?
\end{itemize}

\subsection{The Role of Humans}

The role of the human in the design process will not, and should not, ever be
completely eliminated. The human role may become extremely high level, perhaps
even to the point where we have systems which autonomously infer new
robot designs.  (Imagine, for example, an assistant which notices you performing some
repetitive task and offers a robotic solution.) But humans will still play
a role in the \emph{design of the design tools}, embedding our biases and
preferences. In the more immediate future, our current design technologies rely
heavily on human input for specification design, and for the insights that an
experienced designer can leverage. This expertise is required at all levels of
the robot design stack.

As in all creative fields, robotics has a plethora of design tools for different
types of users and at different levels of abstraction. As research into
automated robot design continues, we must objectively study the effects of different interface and architecture
decisions on how these tools are used, instead of relying on our intuition. For
example, the literature is mixed on whether visual programming languages are
easier to use for novice programmers, despite a widespread belief that they are
better for children and other novices \cite{green1992visual,price2015comparing}.

\begin{figure}
\centering
\includegraphics[scale=0.67]{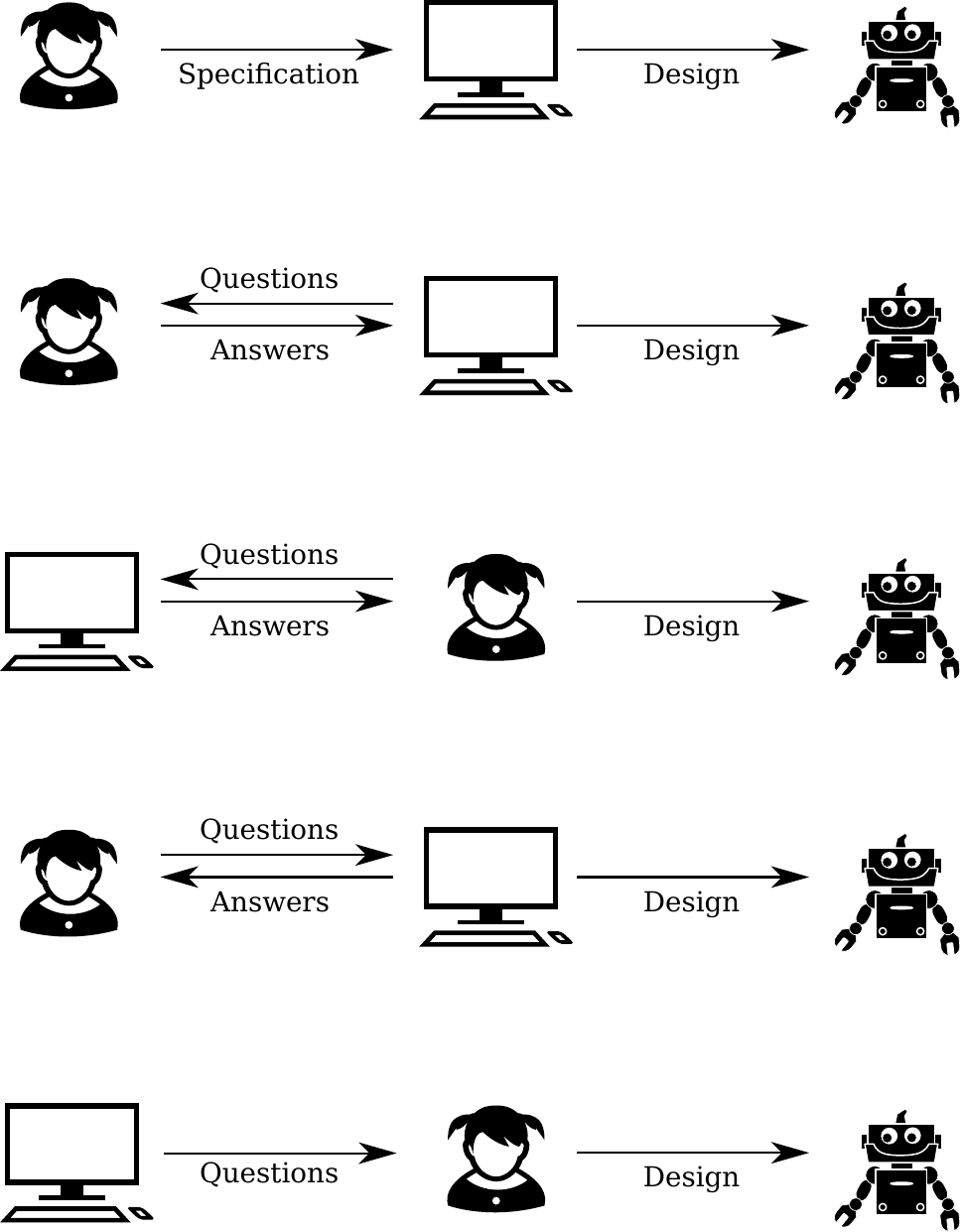}
\caption{A variety of information flows have been proposed for the way computer-aided 
design of robots might proceed. They are distinguished by where the
representation of the robot design is stored (in a computer or in a human
brain), and by the direction and effect of queries.
[First]~A model of traditional synthesis.
[Second]~Automated design, where the automated system queries a human to resolve
incompatibilities or underspecified components of the design.
[Third]~Interactive design tools to answer human designer queries, such as,
``what's the lightest sensor that can detect the color blue?". This category may
broad enough to encompass many rapid-prototyping systems - ``if I 3D print this
wheel, will it be strong and light enough for my robot car?"
[Fourth]~Interactive design tools, where human questions inform the formalized
solution. For example, a human may query the system, ``what happens if I make the
legs twice as long?" which will change the current design solution on the
computer.
[Fifth]~Socratic model: an automated system which asks questions and/or provides
suggestions to inspire a human designer.
(This figure combines contributions from Andrea Censi, Ankur Mehta, and
the present authors)\label{fig:design}.}\vspace*{-22pt} \end{figure}

The following are guiding questions for creators of automated design tools. Many of
these questions are inspired by and explored more deeply in the Human-Computer
Interaction literature, which provides a rich resource for creators of design tools.

\begin{itemize}
\item Who are the intended users of the tool? What other groups of people may
find the tool interesting?
\item What part of the user workflow does the tool replace, or
what new workflows does it enable? 
\item How do users interact with the tool?
    \begin{itemize}
    \item Directly (CAD software, programming
    language, etc) or indirectly (3D printer, robot component database, low-level
    instruction set, etc)? Even indirect interactions are important to consider,
    for example, when a user finds their CAD design won't fit on a print bed, or a
    change in a low-level instruction changes what is possible in a high-level interface.
    \item Modality of interaction: graphical or text based? What kind of feedback
    does the user get from the tool, especially when they specify something
    impossible or introduce a bug?
    \end{itemize}
\end{itemize}


Several interesting examples of these interaction modalities have been explored already, including:
\begin{itemize}
\item Interactive (click and drag) design of morphology and gait with immediate
visual feedback, with fabrication blueprints generated after design is finalized~\cite{bern2017interactive}.
\item Formal specification in code, followed by a
compiler which detects possible problems with
specification and suggests changes to user if the specification has
inconsistencies~\cite{tosun2018computer}.
\item Giving ``early meaning'' to partial designs via dynamical simulations, visualization, and haptic interactions with simulated components~\cite{aukes2014analytic}, and allowing composition of these modules, such as those used in popupCAD~\cite{aukes2015popupcad}.
\end{itemize}

As a baseline, we would like to automate repetitive and time-consuming tasks, and
leave the more creative parts of the workflow intact for the human designer. The ideal case is that a
new automated design tool would enable new forms of human creativity, such as
the way electronic music tools have enabled new methods of human-driven composition
\cite{miranda2001composing}.

\section{Conclusion}
In this abstract, we presented a broad vision for a future in which the process of designing robots transitions from a laborious, error-ridden process driven almost exclusively the cleverness and determination of expert human designers, to one in which automated design tools play a significant role in the process. Our position is that the abstractions commonly used within robotics will have to be extended in order to address questions that are essential for automating hardware realization and fabrication, questions dealing most fundamentally with information and representation.

\bibliographystyle{IEEEtran}
\bibliography{bibworkshop}

\end{document}